\RequirePackage{fix-cm}
\documentclass[smallcondensed]{svjour3} 

\smartqed 
\usepackage{adjustbox}
\usepackage{graphicx}
\usepackage{amssymb}
\usepackage{amsmath}
\usepackage{txfonts}
\usepackage{mathdots}
\usepackage[classicReIm]{kpfonts}
\usepackage{cite}
\usepackage{color}
\usepackage{multirow}
\usepackage{longtable}
\usepackage{ragged2e}
\usepackage[utf8]{inputenc}
\usepackage[english]{babel}
\usepackage{changepage}
\usepackage{fancyhdr}
\usepackage{enumitem}
\usepackage{array}
\usepackage{mathptmx} 
\usepackage{latexsym}
\usepackage{makeidx}  

\begin{document}

\title{ECO-AMLP: A Decision Support System using an Enhanced Class Outlier with Automatic Multilayer Perceptron for Diabetes Prediction}
\titlerunning{ECO-AMLP: A Diabetes Prediction Framework}        

\author{ Maham Jahangir \and Hammad Afzal \and Mehreen Ahmed \and Khawar Khurshid \and Raheel Nawaz}

\institute{Maham Jahangir\at National University of Sciences \& Technology, Islamabad, Pakistan. \\
            \email{mahamjahangir@yahoo.com}             \and
          Hammad Afzal \at National University of Sciences \& Technology, Islamabad, Pakistan. \\
            \email{hammad.afzal@mcs.edu.pk}             \and
          Mehreen Ahmed\at National University of Sciences \& Technology, Islamabad, Pakistan. \\
            \email{mahreenmcs@gmail.com}                \and
          Khawar Khurshid \at National University of Sciences \& Technology, Islamabad, Pakistan. \\
            \email{khawar.khurshid@seecs.edu.pk}        \and
          Raheel Nawaz \at Manchester Metropolitan~University, UK \\
            \email{r.nawaz@mmu.ac.uk}
          }

\maketitle

\begin{abstract}
With advanced data analytical techniques, efforts for more accurate decision support systems for disease prediction are on rise. Surveys by World Health Organization (WHO) indicate a great increase in number of diabetic patients and related deaths each year. Early diagnosis of diabetes is a major concern among researchers and practitioners. The paper presents an application of \textit{Automatic Multilayer Perceptron }which\textit{ }is combined with an outlier detection method \textit{Enhanced Class Outlier Detection using distance based algorithm }to create a prediction framework named as Enhanced Class Outlier with Automatic Multi layer Perceptron (ECO-AMLP). A series of experiments are performed on publicly available Pima Indian Diabetes Dataset to compare ECO-AMLP with other individual classifiers as well as ensemble based methods. The outlier technique used in our framework gave better results as compared to other pre-processing and classification techniques. Finally, the results are compared with other state-of-the-art methods reported in literature for diabetes prediction on PIDD and achieved accuracy of 88.7\% bests all other reported studies.


\end{abstract}

\section{Introduction}
Medical Expert Systems is an active area of research where data analysts and medical experts are continuously striving to make them more accurate using pattern recognition and classification methods. Machine learning algorithms have improved diagnostic systems that help to minimize the cost of conducting extensive medical tests. The improved diagnostic systems with better performance save time of the medical practitioners. Moreover, these systems assist doctors and physicians in their clinical routine. Machine learning algorithms have successfully been applied for diagnosis of various diseases like heart, diabetes, cancer, hepatitis etc \cite{yoo2012data,srinivas2010applications,anbarasi2010enhanced,delen2005predicting,kharya2012using,sathyadevi2011application,tafa2015intelligent}. Particularly, during last few decades, diabetes has become very common which causes an increase in blood glucose level in a person. According to recent statistics by World Health Organization (WHO), 422 million adults have diabetes and 1.5 million deaths are directly attributed to diabetes each year.\footnote{$  $http://www.who.int/diabetes/en/} Therefore, there is an immense need for supporting the medical decision-making process so that diabetes can be detected at an early stage.
\\
A number of predictive frameworks using various classification techniques such as Support Vector Machine (SVM), Artificial Neural Network (ANN), Na\"{i}ve Bayes (NB), Decision Trees (DT) and others are reported in literature\cite{yoo2012data,srinivas2010applications,temurtas2009comparative,nnamoko2014meta}. A systematic literature review revealed that ANNs are the best performer in terms of accuracy of results as compared to other techniques. Various architectures of ANNs have been employed by various researchers in different medical diagnosis \cite{apolloni1990diagnosis,bounds1988multilayer,ohno1997sequential}as ANNs proved to be more flexible in modeling and gives reasonable results in accuracy prediction \cite{park2001sequential}\cite{shanker1996using}. However, one of the major issues with ANNs is that their optimum performance can be achieved using parameter optimization that involves selecting number of hidden layers, neurons, number of epochs and learning rate while defining the network topology of neural network. These parameters are to be decided before training the ANN. This problem is solved by AutoMLP which is a small ensemble of multilayer perceptrons (MLPs) and is auto tunable. It adjusts the parameters automatically.
\\
This paper proposes a novel decision support framework that combines pre-processing techniques with AutoMLP to provide a hybrid prediction model. The proposed system uses an Enhanced Outlier Detection using Distance Based Class Outlier factor as pre-processing of dataset. This removes the outliers from the dataset, which is then fed into Ensemble of MLPs, i.e. AutoMLP. The framework is named as ECO-AMLP: Enhanced class outlier detection combined with AutoMLP. The class outlier factor is determined using probability, deviation and distance of a particular instance with respect to the class label of its K nearest neighbors. The experiments are conducted on publicly available dataset Pima Indians Diabetes Dataset which is used as benchmark dataset in order to compare our technique with existing state-of-the-art approaches. A preliminary study on this framework is provided in \cite{janangir2017diabetes}. A series of experiments are conducted to evaluate the proposed framework where ECO-AMLP is compared with other individual as well as ensemble classifiers. In order to validate the effectiveness of pre-processing technique embedded in ECO-AMLP, experiments are conducted to compare it with other pre-processing techniques such as feature selection, attribute weight generation, normalization, sampling etc. The results demonstrate that the proposed ECO-AMLP outperformed other reported techniques and realized the highest accuracy of 88.7\%. This can be very useful in medical expert systems for which the practitioners and researchers are continuously striving to make them more accurate.
\\
The structure of the paper is as follows. The systematic review of literature is presented in Section 2 that summarizes existing studies in diabetes prediction and description of dataset used. Section 3 provides the proposed framework, followed by the explanation of experiments in Section 4 along with discussion on results. Conclusions and future implications are discussed in Sections 5.

\section{Literature Review}
This section presents a number of studies that employ machine learning techniques in design of intelligent healthcare applications, particularly for prediction of diabetes. We have primarily focused on studies that use pre-processing techniques before applying the learners as they closely resemble our proposed technique. The literature survey conducted during research reported in this paper show that Pima Indians Diabetes Dataset\footnote{$  $http://archive.ics.uci.edu/ml/} (PIDD) is the most commonly used dataset for research related to decision support systems in diabetes prediction. This is a benchmark dataset, commonly used to compare the prediction models. There are other studies reported as well that use privately created datasets, however, the prediction models applied on private datasets cannot directly be compared due to unavailability of these datasets. Therefore, our main focus has been on publicly available dataset PIDD. In terms of mostly used learning techniques, ANN is the most popular prediction model followed by Ensemble-based methods \cite{ahmed2017mcs,ahmed2017improving}. SVM and DTs are also reported to produce good results. A comparative statistics showing the number of studies (reviewed during our research work) using individual and ensemble based classifiers is illustrated in Figure \ref{fig:framework}.
\\

\begin{figure}	
\centering
    \includegraphics*[width=4.47in, height=1.86in, keepaspectratio=false]{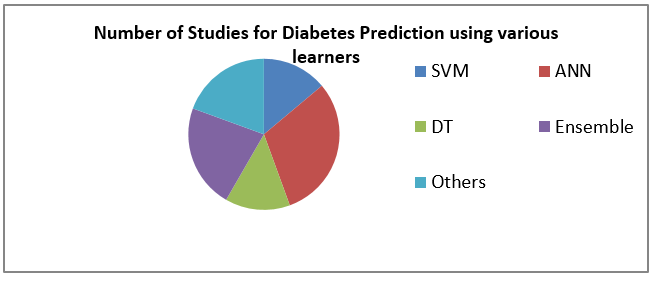}
	\caption{Number of publications (reviewed during this work) related to diabetes prediction using various machine learning methods}
	\label{fig:framework}
\end{figure}

Following text provides a brief description of the dataset PIDD, followed by an overview of existing state-of-the-art learning techniques reported in diabetes prediction. The studies are organized according to the machine learning techniques used in these studies. ANN and Ensemble based techniques are summarized followed by other techniques comprising SVM, DTs and Fuzzy based systems.

\subsection{Pima Indian Diabetes Dataset}
Pima Indians Diabetes Dataset (PIDD) is available on UCI\footnote{$ $http://archive.ics.uci.edu/ml/} machine learning repository. PIDD contains the records of females of at least 21 years of age from the Pima Indian heritage. The number of instances, number of attributes, prevalence of diabetes, and features are listed in Table \ref{tab:pima}.

\begin{table}[]
\centering
    \caption{Description of Pima Indian Diabetes Datasets}
    \label{tab:pima}
\begin{tabular}{|p{0.5in}|p{0.6in}|p{0.6in}|p{0.7in}|p{1.9in}|}\hline
\textbf{Data Set} & \textbf{No. of Instances} & \textbf{No. of Attributes} & \textbf{Prevalence of diabetes} & \textbf{Features}\\ \hline

UCI (Pima Indians) & 768 & 8 & 34.89\% & Number of times pregnant, Plasma glucose concentration, 2 hours in an oral glucose tolerance test, Diastolic blood pressure (mm Hg), Triceps skin fold thickness (mm), 2-Hour serum insulin (mu U/ml), Body mass index, Diabetes pedigree function, Age (years)\\ \hline
\end{tabular}
\end{table}

\subsection{Diabetes Prediction using Artificial Neural Network}
ANNs have been widely used for prediction of diseases \cite{dreiseitl2002logistic,sumathy2010diagnosis,degroff2001artificial,floyd1994prediction,saha2006prediction}. In a survey work related to medical domain, \cite{dreiseitl2002logistic} discussed 72 papers based on ANNs and Logistic Regression (LR) on medical datasets. According to their study, ANN and its variants are the most popular learners while creating decision support frameworks in medical domain. One of the earlier works on diabetes prediction is reported in 2003 who trained different types of ANNs \cite{kayaer2003medical} on PIDD and performed comparative analysis among multilayer perceptron (MLP), Radial Basis Function (RBF) and General Regression Neural Network (GRNN) \cite{specht1991general}. GRNN outperformed other networks by achieving a highest accuracy of 80.2\% on test data. In 2005,\cite{farhanah2005diabetes} applied ANN on the PIDD and used 600 (78\% approx.) randomly selected cases for training set and 168 (22\% approx) for test set. Two different experiments are carried out one with 8 input variables and the other with 4 input variables. They reported highest performance using 8 inputs with 3 hidden layers, showing the correlation coefficient of 1. In 2006,\cite{wettayaprasit2006linguistic} proposed a method of linguistic rule extraction from nodes of ANN and tested it on several UCI benchmark datasets including PIDD. The rules in this paper are extracted from neural network pruning using frequency interval data representation. The if-then rules for diabetes dataset were not mentioned in the paper. They reported an accuracy of 74\% on the PIDD. \cite{salami2010application} suggested the application of CVNN (complex valued neural network) and RVNN (real valued neural network) to PIDD for the prediction of diabetes. The normalization techniques used are z-, min-max, complex and unitary data normalization. Complex data normalization is used to convert Real Value Data to Complex Value Data and ANN based autoregressive model classification technique using different activation functions, learning rate, number of neurons in the hidden layer and the number of epoch. The accuracy ranged from 80\% to 81\% when model was tested using different parameters. The authors further extended their work in \cite{aibinu2011novel} and proposed the application of CVNN combined with complex-valued pseudo autoregressive (CAR) using split weights and adaptive coefficients; thus forming CVNN-based CAR model. The CAR coefficients are obtained from the weights and adaptive coefficients of a trained network. They reported an accuracy of 81.28\% on PIDD. In another study (2009) \cite{temurtas2009comparative} reported an accuracy of 82.37\% using Levenberg--Marquardt (LM)\cite{specht1991general}algorithm with probabilistic neural network on PIDD. Multilayer neural network is trained using LM algorithm. In another work on diabetes prediction, \cite{karan2012diagnosing} proposed an ANN based model and achieved a value of \textbf{1.142} for total square error using 5000 iterations, learning rate 0.95 and momentum of 0.05 on PIDD. They did not report the accuracy of their system.
\\
Diabetes prediction has also been performed using private datasets. Among earlier studies, \cite{park2001sequential} proposed an application of sequential multi layered perceptron (SMLP) using a dataset collected from a US company. Stratified random sampling and random shuffling of inputs is used as pre-processing steps to achieve a sensitivity of 86.04\% and gain (average profit 0.18). In 2006, \cite{pobi2006predicting} performed experiments on Juvenile Diabetes Dataset for prediction and reported an accuracy of 99.72\% using ANN. In the same year, \cite{venkatesan2006application} applied RBF on a private dataset and reported an accuracy of 97.0\% followed by sensitivity 97.3\% and specificity 96.8\%. In 2008, \cite{dey2008application}developed a model based on ANN and applied it on a private dataset collected from Sikkim Manipal Institute of Medical Sciences. Data is normalized as a pre-processing step and ANN is trained using back propagation algorithm whereas gradient descent algorithm is used for updating weights. Network performance of 92.5\% is reported by the authors.

\subsection{Diabetes Prediction using Ensemble Based Learners}
Ensemble-based classification has emerged as a popular technique during last few years in the field of medical diagnostics. The ensemble-based classifiers as explained by \cite{polikar2006ensemble}; is the idea of using a combination of individual classifiers in order to get a classifier that performs better than any of the individual classifier. In 2014, \cite{nnamoko2014meta} proposed a meta model combination of individual classifiers to improve accuracy of diabetes prediction on PIDD. They used Synthetic Minority Over-Sampling Technique (SMOTE) in pre-processing stage to increase the minority class. The dataset is trained on five different learning algorithms: Sequential Minimal Optimization (SMO), Radial Basis Function (RBF), C4.5, NB and RIPPER.  C4.5 produced highest accuracy of 77.9\%, whereas RBF gave the lowest accuracy of 73.6\%. In the next step, they trained a meta model (combiner) of best individual classifier with simple LR algorithm and reported an accuracy of 77.0\% and aROC of 84.9\%. In same year, \cite{li2014diagnosis} used weight adjusted voting approach while training and used an ensemble of SVM, ANN and NB to predict diabetes on PIDD. The records with biologically impossible values are removed during pre-processing and Wrapper method \cite{kohavi1997wrappers} is used for feature selection. They reported an accuracy of 77.0\%, specificity 86.8\% and sensitivity 58.3\%.
\\
Among the studies using Ensemble based classifiers on private datasets, \cite{sabariah2014early} proposed the combination of RF and CART on a dataset collected from medical records of chronic disease of patients from Banjarnegara. They used different number of trees and candidate attributes splitter to get the optimal results. Moreover they analyzed and reported the most relevant attributes in predicting the disease. The findings of this research after series of experiments are: 50 numbers of trees and 3 attributes splitter attained 83.8\% average accuracy. The attributes: \textit{heredity}, \textit{age}, and \textit{body mass index} are regarded as most important and relevant attributes\textbf{. }Another ensemble-based research is reported in \cite{ali2014prediction} in which, authors conducted experiments to classify the dataset into the respective type of diabetes. Adaboost M1 algorithm incorporated with Random Committee is used on a private dataset. RT is used as a base classifier in Random Committee. The algorithm repeatedly runs random tree over various distribution of training diabetes data and combines the outputs in a single random committee classifier. The final output is the average of the results generated by individual random tree classifiers. The authors reported an accuracy of 81.0\% using 10-fold cross validation. In 2015 \cite{han2015rule} applied an ensemble of SVM and RF on dataset collected for China Health and Nutrition Survey (CHNS). The training set is first trained on SVM by tuning parameters to get the highest accuracy, followed by extraction of rules using RF by tuning the rule induction parameters to get the best rules. These rules are then used to predict the class of each record from test data. Pre-processing techniques used are Vacant Data Exclusion, Noise Data Canceling and Feature Selection. The values for precision, recall and f-value calculated after 10-fold cross validation are 81.8\%, 75.6\% and 0.786 respectively.

\subsection{Diabetes Prediction using other Learners (including SVM and DTs)}
One of the earliest work reported in diabetes prediction in 2002, \cite{raicharoen2002critical} applied critical SVM without kernel function to a number of benchmark datasets. The proposed algorithm is also applied on PIDD where reported accuracy is 82.3\% without any cross validation on the PIDD dataset. In 2008, \cite{polat2008cascade} presented the application of Generalized Discriminant Analysis (GDA) \cite{baudat2000generalized} and Least Square SVM (LS-SVM) \cite{suykens1999least} to predict diabetes using PIDD. GDA is used as a pre-processing step, followed by LS-SVM for classification. They reported the accuracy, sensitivity and specificity at 79.16\%, 83.3\% and 82.05\% respectively using 10-fold cross validation. In 2013, \cite{kumari2013classification} applied SVM with Radial Basis Kernel function on PIDD and reported an accuracy of 78\%, 80\% sensitivity and 76.5\% specificity.
\\
Decision tree (DT) and its variants have also been extensively used in diabetes diagnostic. A maximum of 81\% accuracy is reported in studies on PIDD. \cite{han2008diabetes} proposed an application of various data preprocessing techniques combined with decision trees for classification to predict diabetes using PIDD.\textbf{ }The pre-processing techniques used are feature identification and categorization, outlier removal and feature selection, data normalization, numerical data discretization. They reported maximum accuracy of 80\% using ID3. In another study in 2011, \cite{al2011decision} applied DT on PIDD, combining it with attribute identification and selection, handling missing values, and numerical discretization as pre-processing. The dataset is trained using J48 algorithm using 10 fold cross validation and reported an accuracy of 78.2\%. In a recent study, \cite{saxena2015diabetes} achieved an accuracy up to 81.3\% using rules extracted from C4.5 on PIDD.
\\
For completeness, studies using private datasets are reported here; however, they cannot be used for comparison of results as the datasets are not publicly available One such recent study is reported by Tafa et al. \cite{tafa2015intelligent} in 2015 proposed a joint implementation of SVM and Na\"{i}ve Bayes (SVM) on \textbf{Kosovo Diabetes Dataset. }The split ratio for training and test set used is 50:50\%. SVM and ANN are individually trained on the training set. Predictions are based on majority voting from the outputs of both classifiers. The accuracies of 95.52 \% and 94.52\% are reported for SVM and NB respectively. In a similar study, Tama et al. \cite{tama2013early} performed a series of experiments to predict diabetes on a private dataset. SVM outperformed other classifiers as well as ensemble based methods. An average accuracy of \textbf{96.49\%} using hold out and 10-fold cross validation was reported.
\\
A summary of state-of-the-art techniques applied on PIDD and accuracies reported in literature are presented Table \ref{tab:survey}.

\newcolumntype{P}[1]{>{\centering\arraybackslash}p{#1}}
\begin{longtable}{|P{0.2in}|P{1in}|P{1.8in}|P{0.5in}|P{0.5in}|}
    \caption{Summary of selective techniques applied on Pima Indians Diabetes Dataset and their performances}\\ \hline
    \label{tab:survey}
\multirow{2}{1in}{\textbf{Ref \newline Year}} & \multirow{2}{1in}{\textbf{Pre-processing Technique}} & \multirow{2}{1in}{\textbf{Prediction Technique}} & \multicolumn{2}{|P{1in}|}{\textbf{Performance}} \\ \cline{4-5}
& & &  \textbf{Accuracy\newline (\%)} & \textbf{Other Metrics (if available)}\\ \hline
\multicolumn{5}{|P{\linewidth}|}{\centering\textbf{ANN Based Techniques}}\\ \hline
\endfirsthead

\cite{kayaer2003medical}\newline \textbf{2003} & None & General Regression Neural Network (GRNN) & 80.21 & NA \\ \hline
\cite{farhanah2005diabetes}\newline \textbf{2005} & None & ANN & NA & Correlation Coefficient =1 \\ \hline
\cite{wettayaprasit2006linguistic}\newline \textbf{2006} & None & ANN & 74 &  \\ \hline
\cite{temurtas2009comparative}\newline \textbf{2009} & None & Levenberg--Marquardt (LM) algorithm with probabilistic neural network & 82.37 &  \\ \hline
\cite{salami2010application}\newline \textbf{2010} & \textbf{ }Normalization,\newline  Formatting of data & Complex Valued Neural Network (CVNN) & 80-81 &  \\\hline
\cite{aibinu2011novel}\newline \textbf{2011} & None & CVNN [24] based CAR model & 81.28 &  \\ \hline

\multicolumn{5}{|p{\linewidth}|}{\centering\textbf{Ensemble Based Techniques}} \\ \hline
\cite{nnamoko2014meta}\newline \textbf{2014} & Synthetic Minority Over-Sampling Technique (SMOTE) & Meta model of 5 classifier  & 77 & aROC=84.9 \\ \hline
\cite{li2014diagnosis}\newline \textbf{2014} & Missing value imputation,\newline Wrapper method for feature selection & Majority Voting based ensemble method of (SVM + ANN + Na\"{i}ve Bayes) & 77 & Specificity=\newline 86.8\newline Sensitivity=\newline 58.3 \\ \hline

\multicolumn{5}{|P{\linewidth}|}{\centering\textbf{Other Techniques}} \\ \hline
\cite{raicharoen2002critical}\newline \textbf{2002} & None & SVM & 82.29 & NA \\ \hline
\cite{wang2007ontology}\newline \textbf{2007} & None & Ontology Based Fuzzy Inference Agent System  & 74.2 & Precision=\newline 59.2\newline Recall=\newline 71.5 \\ \hline
\cite{han2008diabetes}\newline \textbf{2008} & \textbf{ }Feature identification \& categorization, outlier \newline removal and feature selection,\newline  data normalization, \newline  numerical data discretization & ID3 & 80 &  \\ \hline
\cite{polat2008cascade}\newline \textbf{2008} & None & Generalized Discriminant Analysis (GDA) + Least Square-Support Vector Machine (LS-SVM) & 79.16 & Sensitivity=\newline 83.3\newline Specificity=\newline 82.1 \\ \hline
\cite{al2011decision}\newline \textbf{2011} & \textbf{ }Attribute identification and selection, handling \newline missing values, \newline  numerical discretization & J48 & 78.17 &  \\ \hline
\cite{guo2012using}\newline \textbf{2012} & \textbf{ }Normalization, \newline  Discretization \newline  Feature selection & Na\"{i}ve Bayes Network & 72.3 &  \\ \hline
\cite{daho2013recognition}\newline \textbf{2013} & None & Neuro-fuzzy Classifier & 82.3 & Specificity=\newline 84.6 \newline Sensitivity=\newline 80.8 \\ \hline
\cite{kumari2013classification}\newline \textbf{2013} & None & SVM with RBF Kernel & 78 & Sensitivity=\newline 80 \newline Specificity=\newline 76.5\newline  \\ \hline
\cite{kalaiselvi2014new}\newline \textbf{2014} & None & Neuro-Fuzzy Inference System & 80 &  \\ \hline
\cite{saxena2015diabetes}\newline \textbf{2015} & None & C4.5 & 81.3 &  \\ \hline
\end{longtable}

\section{Proposed Framework: Enhanced Class Outlier Combined with Automatic Multilayer Perceptron}
\noindent The proposed framework amalgamates a number of techniques to provide a hybrid framework that is capable of producing best results on benchmark datasets of diabetes, PIDD. Our framework uses AutoMLP which automatically optimizes the parameters in the architecture of MLP's and minimizes human intervention. The proposed framework improves the performance by employing an enhanced distance based class outlier detection pre-processing technique. The method starts with splitting of dataset into Training, Validation and Test Set. The training dataset is processed initially using a number of pre-processing steps that involve \textit{Data Transformation }(conversion of nominal data into numeric), followed by application of an enhanced class outlier detection that removes noisy and unimportant incidences from datasets. This method detects outliers based on class outlier factor. Using this method, top 10 outliers are detected from the training set based on the 12 nearest neighbors and correlation similarity based distance measure. The outliers detected from the training set are removed which leaves us with a subset of original training set which is outlier free. This dataset is then used for training classifier. In the next step, a variant of Multilayer Perceptron named as AutoMLP is applied that trains on the patients dataset to create a model for prediction of diabetes. AutoMLP performs auto-tuning of parameters, thus providing best parameters (learning rate, hidden layers etc) of MLP as final learner. In this step, an ensemble of 4 MLPs with different number of hidden units and learning rates are used. After ten training cycles the error rate is determined and worst MLP's are trained with the best ones. A complete block diagram of the proposed model is shown in Figure \ref{fig:framework}. Each step involved in framework is described in detail in following sub-sections.

\begin{figure}	
\centering
    \includegraphics*[width=4.23in, height=4.29in, keepaspectratio=false]{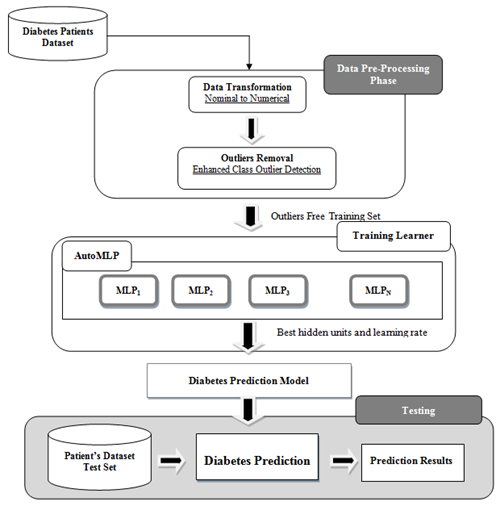}
	\caption{ECO-AMLP Framework for Prediction of Diabetes from Patients Dataset}
	\label{fig:framework}
\end{figure}

\subsection{Data Pre-Processing}
\noindent Each record in PIDD dataset contains the information about a patient (as discussion in Section 2) and a class label with two values; either the patient is healthy or a sufferer of diabetes. The dataset is initially divided into Training, Validation and Test sets comprising of records respectively. Each subset is initially pre-processed to transform the nominal attributes into numeric. The training set is used for further processing in order to produce the prediction model. The process starts with the outlier removals for which we have used Enhanced Class Outlier Distance based algorithm (ECODB). The ECODB is proved to produce better results than other pre-processing techniques).\\

\noindent \textbf{Outlier Detection and Removal: }
Outliers are defined as data points or data instances that are rare cases, exceptions, deviate in behavior from other data points. Conventional techniques detect outliers in data irrespective of the class label i.e. the rare events or exceptions are detected with respect to whole dataset. On the other hand, in \textit{Class Outlier Mining}, the class label is taken into account while detecting outliers in the dataset. In our experiments, an Enhanced Class Outlier Distance Based (ECODB) algorithm is used which is an enhanced version of Class Outlier Distance Based (CODB) algorithm. Both CODB and ECODB have been compared with conventional techniques using a number of public datasets available in \cite{saad2009comparative,hewahi2007class}.\\

\emph{Class Outlier Distance Based (CODB) algorithm}, originally introduced in \cite{hewahi2007class}, detects outliers based on nearest neighbors and distance based approach. It detects outliers based on class outlier factor (COF) which depicts the degree of being a class outlier for a particular data instance. COF considers following key factors: probability of class label of instance among its neighbors, deviation of the particular instance in terms of distances from the respective instances of the same class, distance between the particular instance and its k nearest neighbors. The instance is labeled as class outlier that produces least K-Distance from its K nearest neighbors, its deviation from the respective instances of the same class is the greatest and it has different class label of its K nearest neighbors' class. The COF for any instance (I) as per CODB concept can be represented as: \cite{hewahi2007class}\\

\begin{equation}
    \label{eq1}
    COF\left(I\right)=K\ \times PCL\left(I,K\right)+\alpha \ \times \frac{1}{Deviation\left(I\right)}+\beta \ \times KDist\left(I\right)
\end{equation}

where,
 $PCL\left(I,K\right)\ $ denotes the probability of the class label of the instance (\textit{I)} among the class labels of its \textit{K }Nearest Neighbors. $Deviation\left(I\right)$  is deviation of the instance (I) from the instances of the same class; computed by summing the distances between the instance \textit{(I) }and others. $KDist\left(I\right)$ is summation of distance between (I) and its \textit{K }nearest neighbors. $\alphaup$ and $\betaup$ are controlling factors that control the effects of $Deviation\left(I\right)$\textit{ }and $KDist\left(I\right)$\. Their values are\textit{ }determined by trial and error. CODB rank each instance based on the above formula \eqref{eq1} and detect the outliers in the dataset.\\

\emph{Enhanced Class Outlier Distance Based algorithm} is an enhanced version of CODB proposed by same authors \cite{saad2009comparative}. It gets rid of the hit and trial method for adjusting values of $\alphaup$ and $\betaup$. The ECODB algorithm defines Class Outlier Factor of particular instance $COF\left(I\right)$ as:

\begin{equation}
    \label{eq2}
     ECOF(I)=K \times PCL(I,K)-norm(Deviation(I))+norm(KDist(I))
\end{equation}

ECODB uses normalized values of $Deviation\left(I\right)$\textit{ }and $KDist\left(I\right)$ instead of $\alphaup$  and $\betaup$ with a value range of [0-1]

\begin{equation}
    \label{eq3}
    norm\left(Deviation\left(I\right)\right)=\frac{Deviation\left(I\right)-MinDev}{MaxDev-MinDev}
\end{equation}

\begin{equation}
    \label{eq4}
    norm\left(KDist\left(I\right)\right)=\frac{KDist\left(I\right)-MinKDist}{MaxKDist-MinKDist}
\end{equation}

where, $MaxDev$ and $MinDev$ represent the highest and lowest deviation value for top N class outliers. Similarly, $MinKDist$ and $MaxKDist$ are lowest and highest $KDist$ value for top N class outliers. A number of experiments are performed using various variants of ECODB by considering different number of neighbors, different number of class outliers, measure types (numerical, mixed, nominal) and numerical measure. Measuring type is used for selecting the type of measure to be used for measuring the distance between points, e.g. Euclidean Distance, Cosine Based Similarity etc. As previously discussed, ECODB provides the instances which deviate the most from normal trend in dataset. An N number of outliers can be selected by repeatedly performing the algorithm. We experimented with various numbers of outliers to measure the performance of training process. The best results are achieved using 12 number of neighbors, 10 number of class outliers and numerical measure type as correlation similarity. Therefore, these 10 instances are detected and removed from the training set to perform further steps.

\subsection{Training the Learner}
For training process, our framework utilizes the strength of AutoMLP and creates an ensemble of AutoMLPs. The topology of network while designing the ANNs is of utmost significance. ANN, like human brain, contains of a network of inter-connected neurons where each connection has associated weight with it. These weights are adjusted based on the learning experience of the algorithm. The network topology for ANNs has to be adjusted before training the algorithm that includes the number of hidden layers and hidden units in them, learning rate (training parameter that controls the size of weight and bias changes), number of epochs (number of iterations overtraining set) have to be adjusted to train the network. Parameter Optimization is a long run problem of ANNs \cite{rashid2012scanning} which required human intervention to choose the best suitable parameters for the network. However, AutoMLP works on a mechanism to optimize the parameters involved in structure of ANN. Working of AutoMLP is briefly described here for completion of reference. The experimental setup has been shown in Figure \ref{fig:framework}.\\

AutoMLP introduced in \cite{breuel2010automlp}, is a type of multi layered feed forward neural network which is auto tunable i.e. it automatically adjusts learning rate and number of hidden units. AutoMLP combines ideas from genetic algorithm and stochastic optimization. It maintains a small ensemble of networks (MLPs) that are trained in parallel with different number of hidden units and learning rate using gradient based optimization techniques. The error rate is determined on a validation set after a small fixed number of epochs followed by replacing worst performer networks with best ones. This way the networks have different number of hidden units and learning rates. Learning rates and hidden unit numbers are drawn according to probability distributions derived from successful rates and sizes. In PIDD dataset, the instances comprising of attributes such as plasma glucose concentration, BMI, diabetes pedigree function etc are fed as input to the architecture and constitute the input layer. The numbers of attributes for PIDD are 7. These inputs are weighted and then pass from input layer to hidden layer. MLP applies a non-linear activation function to the weighted input. The parameters provided to AutoMLP for training are:\\

\begin{enumerate}
\item  \textbf{Training cycles:} The number of maximum training cycles used for the neural network training.
\item  \textbf{Number of generations:} The number of generations for AutoMLP training.
\item  \textbf{Number of Ensemble MLPs:} The number of MLPs per ensemble.
\end{enumerate}

Experiments are performed by varying the number of these three parameters. After pre-processing and training, the performance of the classifier is evaluated using validation and test set.\\

\paragraph{Experimental Setup}
The proposed framework combines ECODB with AutoMLP as shown in Figure II.  The figure illustrates that the original medical dataset is first subjected to data transformation in which any nominal attribute data is changed to numerical data. After data transformation the training set is fed to outlier detection phase. These outliers are then removed as described earlier and illustrated in the figure. The next step is training the learner. Four MLP's are trained in proposed setup and the best MLP is replaced with the worst ones after 10 training cycles. The numbers of generations used are also 10. The network topology for the best MLP selected after training process consisted of one hidden layer and 160 nodes in hidden layer. Sigmoid is used as an activation function for adjusting weights at hidden nodes to obtain final result.

\paragraph{Performance Metrics used during Training}
Performance Metrics refer to the evaluators that determine the performance of classifier. It is the quantitative measure of how accurately the classifier predicts or classifies the class variable. In order to tune our ensemble setup, we made use of following metrics: Accuracy, Precision and Recall. These metrics are then further used to evaluate the performance of proposed framework on test data. The metrics are briefly described below.

\begin{enumerate}
    \item  \textbf{True positives (TP)}: TP refers to the number of actual diabetic patients correctly predicted by our framework.
    \item  \textbf{True negatives (TN): }TN refers to the non-diabetic patients predicted as non-diabetic by our framework.
    \item  \textbf{False positives (FP): }FP refers to the non-diabetic patients predicted diabetic by the framework.
    \item  \textbf{False negatives (FN):} FN refers to the actual diabetic patients predicted non-diabetic by the framework.
\end{enumerate}

Using the above stated variables, the evaluation metrics can be defined as following:\\

\textbf{Accuracy} is the percentage of patients that are correctly diagnosed by classifier (diabetic or non-diabetic).
\[\mathrm{Accuracy=}\frac{\mathrm{TP+TN}}{\mathrm{P+N}}\]


\textbf{Precision/Specificity }represents the correctness of diabetic diagnosis i.e. Percentage of patients labeled as diabetic are actually diabetic (exactness)
\[\mathrm{Precision=}\frac{\mathrm{TP}}{\mathrm{TP+FP}}\]

\textbf{Recall/Sensitivity} represents the completeness of coverage, i.e.\textbf{ }Percentage of actual diabetic patients correctly diagnosed by our system\\
\[\mathrm{Recall=}\frac{\mathrm{TP}}{\mathrm{TP+FN}}\]

\textbf{Weighted Mean Precision} is the average of precision obtained per class (two classes)\\
\indent \textbf{Weighted Mean Recall} is the average of recall calculated per class (two classes).\\

\section{Experimental Results and Discussion}
In order to perform experiments, PIDD is divided into Training, Validation and Test sets having 70\%, 15\% and 15\% in each set respectively. The performance is evaluated on the test sets keeping the same parameters as those tuned on validation set. It is ensured, that testing data remains unseen and is not manipulated during experimentation in any way. The numbers of instances in Training, Validation and Test sets are 528, 115 and 115 respectively for PIDD as summarized in Table \ref{tab:statistics}.

\begin{table}[]
    \centering
    \caption{Statistics related to two experimental setups}
    \label{tab:statistics}
\begin{tabular}{|p{1in}|p{0.8in}|p{0.6in}|p{0.6in}|p{0.4in}|} \hline

\textbf{Experiments} & \textbf{Number of Features} & \textbf{Training Set} & \textbf{Validation Set} & \textbf{Test Set}\\ \hline
 & 7 & 528 & 115 & 115 \\ \hline
\end{tabular}
\end{table}

A series of detailed experiments are carried out on PIDD using both setups in order to establish the supremacy of proposed approach as compared to other state-of-the-art pre-processing techniques and learners. The experiments can be generally categorized as following:

\begin{enumerate}
    \item  Performance comparison of used outlier method with other pre-processing techniques (details in Section 4.1).
    \item  Performance comparison of the used Ensemble of AutoMLPs with other classifiers (details in Section 4.2).
    \item  Performance comparison of proposed complete framework: ECO-AMPL (Outliers removal using ECODB combined with Ensemble of AutoMLP) with other state-of-the-art results on benchmark dataset i.e. PIDD (details in Section 4.3).
\end{enumerate}

\subsection{Comparison of ECODB (outlier removal technique used in our framework) with other Pre-Processing Techniques}
The experiments are performed by varying the pre-processing techniques combined with Ensemble of AutoMLP and results are compared with the proposed framework, i.e. ECO-AMLP. In specific, we change only pre-processing techniques and keep rest of the components of our framework fixed in order to verify the better performance of ECODB. The results demonstrate that the hybrid combination used in ECO-AMLP performs better than other methods. The Accuracies, Weighted Mean Recall and Weighted Mean Precision achieved by the proposed framework and other methods on PIDD are presented in Table \ref{tab:res_outlier}. Different feature selection, attribute weight generation normalization and sampling techniques are compared with outlier detection using ECODB.

\begin{table}[]
    \centering
    \caption{Comparison of ECODB outlier detection with other pre-processing techniques used on PIDD. Accuracy, Weighted Mean Precision and Weighted Mean Recall are presented in percentage}
    \label{tab:res_outlier}
\begin{tabular}{|p{2.3in}|p{0.6in}|p{0.3in}|p{0.3in}|} \hline

\textbf{Pre-Processing Methods} & \textbf{Accuracy} & \textbf{WMR} & \textbf{WMP} \\ \hline

\multicolumn{4}{|p{\linewidth}|}{\centering\textbf{Feature Selection}} \\ \hline
Principal Component Analysis & 65.04 & 62.59 & 62.77 \\ \hline
Fast Correlation-based Filter (FCBF)\textbf{} & 82.61 & 74.00 & 82.31 \\ \hline
Select by recursive feature elimination with SVM\textbf{} & 82.61 & 74.00 & 82.31 \\ \hline
Select by Feature Quantile Filter\textbf{} & 82.61 & 78.27 & 79.31 \\ \hline
\multicolumn{4}{|p{\linewidth}|}{\centering\textbf{Attribute Weight Generation}} \\ \hline
Weight by Maximum Relevance\textbf{} & 82.61 & 74.00 & 82.31 \\ \hline
Weight by correlation based weak association\textbf{} & 82.61 & 74.00 & 82.31\newline  \\ \hline
Generate Weight Stratification\textbf{} & 78.26 & 68.36 & 75.61 \\ \hline
Generate Weight (LPR)\textbf{} & 82.61 & 83.39 & 79.43 \\ \hline
\multicolumn{4}{|p{\linewidth}|}{\centering\textbf{Normalization}} \\ \hline
Normalization (Z-Transform)\textbf{} & 29.57 & 50.00 & 14.78 \\ \hline
\multicolumn{4}{|p{\linewidth}|}{\centering\textbf{Sampling}} \\ \hline
Bootstrap Sampling\textbf{} & 81.74 & 75.09 & 79.08 \\ \hline
Stratified Sampling\textbf{} & 81.74 & 75.09 & 79.08 \\ \hline
\textbf{Proposed Framework Outlier Detection using  enhanced class outlier distance based} & \textbf{88.7} & \textbf{88.56} & \textbf{85.83} \\ \hline
\end{tabular}
\end{table}

It is evident from the above shown results that the ECODB combined with the Ensemble of AutoMLP outperformed other pre-processing techniques with more than 5\%.

\subsection{Comparison of Ensemble of AutoMLP (classifier) with other classifiers}
In next phase, experiments are performed by keeping the pre-processing technique fixed i.e. ECODB and varying the classifiers and then results are compared with the proposed framework, i.e. ECO-AMLP. This verifies the performance of AutoMLP. The results shown in Table V demonstrate that the optimal combination used in ECO-AMLP performs better than other methods. As compared to accuracy at 88.70\% of proposed method, the highest accuracy achieved using other methods is 81.74\% with SVM. The lowest accuracy is 74.78\% using KNN. The results of other classifiers as reported by literature are also improved in this research as ECODB proved to be better option for pre-processing the data. For example in \cite{kumari2013classification}, accuracy of SVM is 78\%. Similarly literature reports an accuracy of  78.17\% \cite{al2011decision} using DTs while our framework reached an accuracy of 79.13\%. Comparison is also performed between other flavours of ANN and Ensemble of AutoMLP used. Results are shown in Table \ref{tab:others}.

\begin{figure}	
\centering
    \includegraphics*[width=3in, height=2in, keepaspectratio=false]{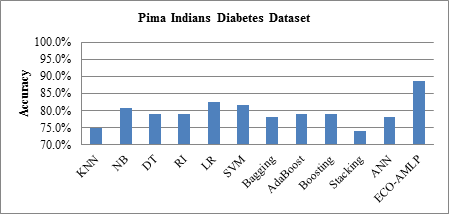}\\
	\caption{Comparison of ECO-AutoMLP with other classifiers}
	\label{fig:others}
\end{figure}

\begin{figure}	
\centering
    \includegraphics*[width=3in, height=2in, keepaspectratio=false]{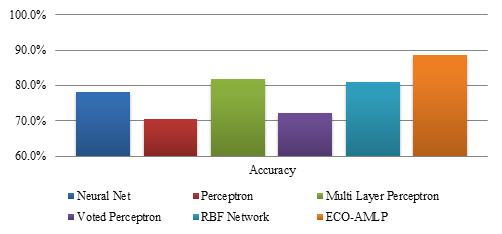}\\
	\caption{Comparison of ECO-AutoMLP with other ANN Based Systems}
	\label{fig:anns}
\end{figure}

\begin{table}[]
    \centering
    \caption{Comparison of proposed framework (ECO-AMLP) with state of art classification techniques on PIDD. Accuracy, Weighted Mean Precision and Weighted Mean Recall are presented in percentage}
    \label{tab:others}
\begin{tabular}{|p{1in}|p{0.5in}|p{0.5in}|p{0.5in}|} \hline
\textbf{Classification Technique} & \textbf{Accuracy} & \textbf{WMR} & \textbf{WMP} \\ \hline
\textbf{KNN} & 74.78 & 71.86 & 70.31 \\ \hline
\textbf{Na\"{i}ve Bayes} & 80.87 & 77.03 & 77.03 \\ \hline
\textbf{Decision Tree} & 79.13 & 68.12 & 78.58 \\ \hline
\textbf{Rule Induction} & 79.13 & 71.53 & 75.63 \\ \hline
\textbf{Linear Regression} & 82.61 & 73.15 & 83.55 \\ \hline
\textbf{SVM} & 81.74 & 75.53 & 81.43 \\ \hline
\textbf{Bagging} & 78.26 & 70.92 & 74.24 \\ \hline
\textbf{AdaBoost} & 79.13 & 74.95 & 74.95 \\ \hline
\textbf{Boosting} & 79.13 & 74.95 & 74.95 \\ \hline
\textbf{Stacking} & 73.91 & 68.68 & 68.68 \\ \hline
\multicolumn{4}{|p{\linewidth}|}{\centering\textbf{Different Architectures of Artificial Neural Network}} \\ \hline
\textbf{Artificial Neural Net} & 78.26\textbf{} & 71.77\textbf{} & 74.04\textbf{} \\ \hline
\textbf{Perceptron} & 70.43 & 52.56 & 60.78 \\ \hline
\textbf{Multi Layer Perceptron} & 81.74 & 75.94 & 78.65 \\ \hline
\textbf{Voted Perceptron} & 72.17 & 58.91 & 65.50 \\ \hline
\textbf{RBF Network} & 80.87 & 74.47 & 77.67 \\ \hline
\textbf{Proposed Technique\newline (ECO-AMLP)\newline } & \textbf{88.70} & \textbf{88.56} & \textbf{85.83} \\ \hline

\end{tabular}
\end{table}

\subsection{Comparison of ECO-AMLP with existing prediction techniques}
\noindent In last, in order to establish the supremacy of propose method over existing state-of-the-art approaches reported for prediction of diabetes, the comparison of proposed technique with the ones reported in literature is presented in Table VI. Various prediction methods comprising different classifiers including ANNs have been employed. The highest accuracies ranging from 81 to 82\% have been reported using ANNs. The pre-processing techniques, prediction technique and the performance evaluators are detailed in the Table \ref{tab:res_literature}. It can be clearly seen that the proposed technique outperformed techniques presented in literature.

\begin{table}[]
    \centering
    \caption{Comparison of proposed technique with literature on Pima Indians Dataset}
    \label{tab:res_literature}
\begin{tabular}{|p{0.5in}|p{1.2in}|p{1.1in}|p{0.5in}|p{1.0in}|} \hline

\multirow{2}{*}{{\textbf{Ref/Year}}} & \multirow{2}{*}{{\textbf{Pre-processing Technique}}} & \multirow{2}{*}{{\textbf{Prediction Technique}}} & \multicolumn{2}{|p{1.5in}|}{\textbf{Performance}} \\
\cline{4-5}

\textbf{} & \textbf{} & \textbf{} & \textbf{Accuracy} & \textbf{Other Metrics (if available)} \\ \hline
\cite{salami2010application}/2010 & Normalization and formatting of data & Complex Valued Neural Network (CVNN) & 81 &  \\ \hline
\cite{aibinu2011novel}/2011 & None & CVNN based CAR model & 81.28 &  \\ \hline
\cite{guo2012using}/2012 & Normalization, Discretization and Feature Selection & Na\"{i}ve Bayes Network & 72.3 &  \\ \hline
\cite{daho2013recognition}/2013 & None & Neuro-fuzzy Classifier & 82.32 & Specificity=84.60\newline Sensitivity=80.76 \\ \hline
\cite{nnamoko2014meta}/2014 & Synthetic Minority Over-Sampling Technique (SMOTE) & Meta model of 5 classifier  & 77.0 & aROC=84.9 \\ \hline
\cite{kalaiselvi2014new}/2014 & None & Neuro-Fuzzy Inference System & 80 &  \\ \hline
\cite{saxena2015diabetes}/2015 & None & C4.5 & 81.27 &  \\ \hline
2016\newline Proposed Technique & \textbf{Outlier Detection using ECODB} & \textbf{Auto MLP} & \textbf{88.70} & \textbf{Weighted Mean Recall=88.56\newline Weighted Mean Precision=85.83} \\ \hline

\end{tabular}
\end{table}

\section{Conclusions}
This paper presents a novel framework Enhanced Class Outlier based algorithm with AutoMLP (ECO-AMLP) to predict diabetes from a public dataset of patients' named as Pima Indian Diabetes Dataset PIDD. Paper summarizes the reported studies on PIDD and other private datasets and presents a number of experiments performed using ECO-AMLP on PIDD to show that the proposed technique provides promising results. Instead of relying on complex feature selection or extraction tasks, paper uses an outlier detection based technique as a pre-processing step to detect and remove outliers in dataset. A series of experiments are performed that show that ECODB performed better as compared to other normalization, attribute weight generation and feature selection techniques. Moreover, ECODB, when applied to other classifiers gave better results than those reported in literature.\\

The systematic literature review revealed that the neural structures can be successfully used to predict diabetes. The selection of optimal number of hidden units and learning rate is a long run problem while defining the network topology for neural network architectures. For this purpose this paper presents the application AutoMLP. The proposed framework used an ensemble of four MLP's to achieve greater accuracy. The AutoMLP gave higher accuracy weighted mean recall and precision when compared with other architectures of neural network. The proposed technique is also compared with the results reported in literature. The experimental results prove that the proposed framework achieved an accuracy of 88.7\% with PIDD which bests the highest reported accuracies.  The most relevant attributes for diabetes prediction in PIDD are: plasma glucose concentration, diastolic blood pressure and number of times pregnant.\\

\bibliography{diabetes}
\bibliographystyle{spmpsci}      
\end{document}